# LLM-based Agents for Automated Confounder Discovery and Subgroup Analysis in Causal Inference

**Po-Han Lee**
*Department of Information Management*
*National Sun Yat-Sen University*
Kaohsiung, Taiwan
jasonleepohan@gmail.com

**Yu-Cheng Lin**
*Department of Information Management*
*National Sun Yat-Sen University*
Kaohsiung, Taiwan
yuchenglinn@gmail.com

**Chan-Tung Ku**
*Department of Information Management*
*National Sun Yat-Sen University*
Kaohsiung, Taiwan
kuchantung@gmail.com

**Chan Hsu**
*Department of Information Management*
*National Sun Yat-Sen University*
Kaohsiung, Taiwan
chanshsu@gmail.com

**Pei-Cing Huang**
*Department of Information Management*
*National Sun Yat-Sen University*
Kaohsiung, Taiwan
pcpeicing@gmail.com

**Ping-Hsun Wu**
*Division of Nephrology*
*Kaohsiung Medical University Hospital*
Kaohsiung, Taiwan
970392@kmuh.org.tw

**Yihuang Kang**
*Department of Information Management*
*National Sun Yat-Sen University*
Kaohsiung, Taiwan
ykang@mis.nsysu.edu.tw

*Abstract*—Estimating individualized treatment effects from observational data presents a persistent challenge due to unmeasured confounding and structural bias. Causal Machine Learning (causal ML) methods, such as causal trees and doubly robust estimators, provide tools for estimating conditional average treatment effects. These methods have limited effectiveness in complex real-world environments due to the presence of latent confounders or those described in unstructured formats. Moreover, reliance on domain experts for confounder identification and rule interpretation introduces high annotation cost and scalability concerns. In this work, we proposed Large Language Model-based agents for automated confounder discovery and subgroup analysis that integrate agents into the causal ML pipeline to simulate domain expertise. Our framework systematically performs subgroup identification and confounding structure discovery by leveraging the reasoning capabilities of LLM-based agents, which reduces human dependency while preserving interpretability. Experiments on real-world medical datasets show that our proposed approach enhances treatment effect estimation robustness by narrowing confidence intervals and uncovering unrecognized confounding biases. Our findings suggest that LLM-based agents offer a promising path toward scalable, trustworthy, and semantically aware causal inference.

## I. INTRODUCTION

Modeling heterogeneous treatment effects (HTE) of interventions using observational data has become a critical challenge in healthcare research, particularly for advancing personalized medicine. While randomized controlled trials (RCTs) remain the gold standard for estimating treatment effects, their cost and time demands limit their widespread adoption and applicability. On the other hand, observational datasets are more readily available and less costly but lack the benefit of randomization, making them vulnerable to confounding biases that can invalidate causal conclusions [1]. Consequently, identifying and appropriately adjusting for confounders to mitigate bias has become a central problem in causal machine learning (causal ML) [2].

Current causal ML methods encompass a wide range of techniques, from interpretable tree-based models to highly flexible neural network architectures [3], [4], [5]. These methods seek to accurately estimate HTE by explicitly identifying confounders and covariates or modeling the outcome response surfaces for treated and untreated groups. While neural networks often achieve superior predictive accuracy, their black-box nature hampers interpretability, making it difficult to verify whether the underlying inference aligns with domain knowledge, an essential requirement in healthcare applications where trust and transparency are paramount. Conversely, interpretable models such as causal trees offer greater transparency but are limited in their capacity to identify relevant confounders comprehensively and to provide stable estimates due to their sensitivity to data variability.

In this paper, we propose a novel AI-in-the-loop framework that balances the trade-off between model interpretability and precise estimation of heterogeneous treatment effects. Our framework constructs a Mixture of Experts (MoE) model [6] composed of causal trees through a two-step iterative process involving confounder verification and uncertainty evaluation. First, during confounder verification, we employ medical Large Language Models (LLMs) [7] as AI agents [8] to screen candidate confounders derived from the partition rules of causal trees. We further incorporate the agents with domain knowledge by Retrieval Augmented Generation (RAG) [9] to enhance the reliability of screening. These AI-suggested confounders are then reviewed and validated by domain experts, facilitating a collaboration that ensures domain-consistent confounder identification while reducing expert workload. Second, in the uncertainty evaluation step, we quantify the estimation uncertainty for each sample and filter out those with high uncertainty. We then train an additional causal tree on these uncertain samples and the remaining covariates to discover potential confounders overlooked in previous iterations. This iterative refinement continues until the uncertainty falls below a predefined threshold. At this point, the framework outputs a final MoE model comprised of validated causal trees

We demonstrate the effectiveness of our framework through a clinical application focused on the unbiased estimation of the HTE in Acute Coronary Syndrome (ACS). Given that the ACS is a leading cause of morbidity and mortality worldwide, accurately assessing the risks and benefits of medications in this context poses significant challenges. Identifying confounding comorbidity can lead to more robust risk assessments and help uncover drug-disease interactions relevant to ACS. Our experimental results indicate that the proposed framework:

1. identifies samples possibly confounded by unobserved confounders;
2. corrects bias introduced by observed confounders, enabling unbiased estimation of HTE;
3. reduces the burden of domain experts in confounder identification.

The remainder of this paper is organized as follows: Section 2 provides the background of our study across causal inference models, LLM-based agents, and RAG. Section 3 details the idea and components in the proposed framework, LLM-based agents for automated confounder discovery and subgroup analysis in causal inference. The experiment designs and results are demonstrated in Section 4. The last section concludes our work and contributions.

II. BACKGROUND

Reliably estimating treatment effects is essential for assessing the effectiveness of treatments or policies across various domains, including healthcare and economics. Achieving robust estimates of treatment effects requires a causal inference framework to appropriately adjust for confounding [10]. In prior research, direct estimators such as Causal Inference Trees [4] and Causal Trees [3] have been proposed to identify confounding variables and manage their impact on treatment effect estimation. However, tree-based models are known to be sensitive to small perturbations in the input data. Even small perturbations in the training dataset can lead to altered tree structures and inconsistent results.

To mitigate this instability, we propose adopting the MoE structure [6]. The MoE architecture employs a Divide-and-Conquer strategy to partition the input space into multiple subregions, enabling each expert to specialize in modeling localized structures. By narrowing the scope of each expert, the overall model becomes less susceptible to bias due to confounding. Moreover, the gating network in the MoE architecture assigns weights to each expert's output. If one expert overreacts to a small perturbation, others may remain unaffected, allowing the gating network to downweight the contribution of the unreliable expert and enhance the robustness of the model.

When integrating the MoE structure with causal tree-based methods, ensuring the resulting approach yields credible and robust estimates of causal effects is essential. Recent research has demonstrated that constructing valid confidence intervals (CIs) for the average treatment effect (ATE) plays a critical role in assessing whether the estimated treatment effects are indeed credible and robust [11]. By quantifying the uncertainty surrounding treatment effect estimates, confidence intervals serve as a statistical tool to validate causal claims. Furthermore, narrowing the width of these confidence intervals has important implications. A smaller CI width indicates greater precision in the treatment effect estimate, which suggests a more reliable and stable causal inference.

Beyond addressing data sensitivity issues, another critical challenge in causal tree methods is their reliance on the assumption that all confounding variables are fully and correctly observed. Causal trees rely on the assumption of unconfoundedness to produce causal estimates. However, the failure to account for unobserved confounders can result in spurious associations and misleading conclusions [12]. As a result, treatment effect estimates derived from causal tree-based approaches are subject to bias and inconsistency. Significant advancements have been made in identifying confounders in structured observational data [13]. Nonetheless, these methods are limited in their ability to process unstructured inputs, such as free-text clinical notes or natural language descriptions [14]. This constraint leads to the underutilization of valuable, context-rich information embedded in unstructured data, highlighting a significant gap between the capabilities of current causal machine learning methods and the complexities inherent in real-world applications.

This gap motivates using LLMs that process unstructured data and may correct the estimation drift. Recently, advances in LLMs have established a new paradigm in developing AI agents, namely LLM-based agents [8]. In agent-based modeling frameworks, each agent represents an autonomous entity whose actions are governed by explicit rules for learning, decision-making, and adaptation. These agents simulate complex task environments by capturing behaviors such as task allocation, information flow, and inter-agent coordination [15]. Unlike standalone LLMs, agent-based architectures are capable of perceiving and interacting with external resources by invoking semantic functions, accessing APIs, and retrieving knowledge from structured or unstructured sources [16]. This shift toward agents enables models to plan, perform reasoning, take actions, and incorporate feedback. A pivotal innovation in the advancement of LLM-based agents is the development of Chain-of-Thought (CoT) prompting [17], which enables models to deconstruct tasks into a sequence of intermediate reasoning steps. As task complexity rises, the effectiveness of single-prompt CoT approaches tends to decline. This decline often leads to reasoning bottlenecks and incomplete solution paths. To address this challenge, Decomposed Prompting [18] has been proposed as an extension of CoT, introducing a modular task decomposition framework. This method explicitly partitions a complex task into manageable sub-tasks, each addressed by a dedicated prompt, specialized model component, or symbolic function.

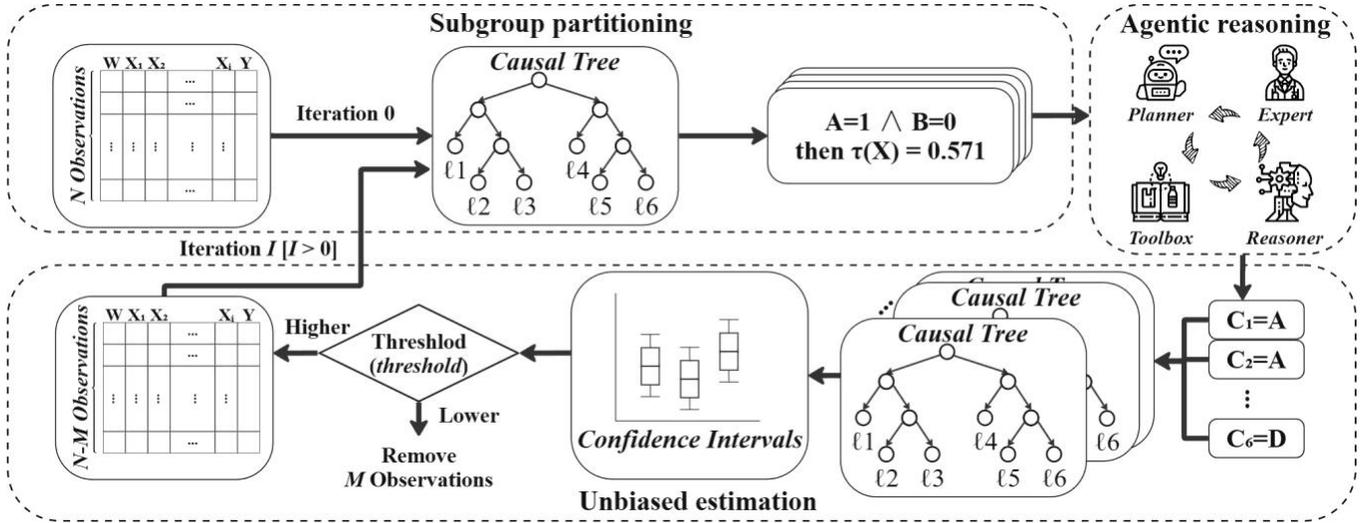

Fig. 1. LLM-based agents for automated confounder discovery and subgroup analysis framework.

Despite their impressive generative capabilities, LLMs face fundamental limitations in knowledge retrieval and manipulation, often underperforming task-specific systems on knowledge-intensive applications. To overcome these constraints, LLM-based agents rely on external tools that provide access to up-to-date and verifiable information. A representative example is RAG [9], which embeds documents into a semantic vector space, enabling agents to perform similarity-based retrieval of contextually relevant information. Retrieved content is dynamically injected into the model's prompt at inference time, grounding responses in external sources and thereby reducing hallucinations while improving factual accuracy and task relevance.

Integrating such capabilities into causal machine learning pipelines offers promising opportunities to enhance inference accuracy and reduce manual effort [19]. Nevertheless, the collaboration between LLM-based agents and causal workflows remains relatively underexplored. Building on agents with promising reasoning abilities, we aim to enhance causal MLs' capabilities for unbiased estimation and reduce human costs..

## III. LLM-BASED AGENTS FOR AUTOMATED CONFOUNDER DISCOVERY AND SUBGROUP ANALYSIS IN CAUSAL INFERENCE

In this section, we present LLM-based agents for automated confounder discovery and subgroup analysis, a framework designed to balance the trade-off between model interpretability and the accurate estimation of heterogeneous treatment effects. Our framework adopts a divide-and-conquer architecture inspired by the MoE design. This allows the model to learn heterogeneous subgroups, each influenced by different confounding factors, as shown in Figure 1. To achieve this, our framework operationalizes its approach through three key steps: (1) partitioning observational data into meaningful subpopulations for initial treatment effect estimation; (2) extracting decision rules and retrieving domain-specific knowledge to support causal reasoning; and (3) refining effect estimates for unstable samples through iterative adjustment. In each iteration, the model focuses on a subset of the data that remains affected by confounding. It refines its focus in successive steps.

### A. Subgroup partitioning

To initiate the causal analysis, we employ a Causal Tree to estimate the initial treatment effects. Given a training sample $S_{tr} = \{Y_i^{obs}, W_i, X_i\}_{i=1}^n$, where $Y_i^{obs}$ is the observed outcome, $W_i$ is a binary treatment indicator, and $X_i$ are pre-treatment variables. The tree is split recursively based on the heterogeneity of treatment to define a set of subgroups $\Pi$, with $\Pi$ the number of subgroups (leaves) in the partition. We write $\Pi = \{\ell_1, \ell_2, \ldots, \ell_j\}$. Each leaf node $\ell_j$ is associated with a CATE, denoted $\tau(X)$. As an example, a single leaf node $\ell_j$ can be represented using a rule-based representation, which relies on symbolic conjunctions (AND) to define the subgroup. These estimated $\tau(X)$ and their partition of the covariate space $\Pi$ form the causal state space for subsequent reasoning, and serve as input to an LLM-based agent.

### B. Agentic reasoning

At the heart of our approach is an LLM-based agent, which simulates expert judgment by generating, decomposing, and validating causal hypotheses about subgroup heterogeneity. While Figure 1 provides an overview of the agent's workflow, the detailed reasoning process is described in Algorithm 1.

The agent will extract rule $R = \{r_1, r_2, \ldots, r_i\}$ from $\Pi$, by explaining the tree structure and the meaning of each $\ell_i$ along with its $\tau(X)$. Each $r_i$ captures both the distributional characteristics of the covariates $X$ within $\ell_i$ and provides a rationale for why the corresponding $\tau(X)$ emerges. Once these rules are extracted and elaborated, the next step involves decomposing them into actionable statuses that guide the agent's reasoning process. We employ a decomposed prompting strategy, where each $r_i$ is broken down into a sequence of subqueries $Q_r = \{q_{r1}, q_{r2}, \ldots, q_{ri}\}$. For instance, $q_r$ may include questions about how specific covariates $X$ influence outcomes $Y$, or how the treatment variable $W$ interacts with $X$ in shaping the causal relationships.

**Algorithm 1** Agentic Causal Reasoning for Confounder Identification
___
1: **Inputt:** $\Pi$, $\tau(X)$
2: **Propose** $R = \{r_1, r_2, \ldots, r_i\}$ by extracting and explaining the meaning of features from $\Pi$ and $\tau(X)$
3: **Propose** $Q_r = \{q_{r1}, q_{r2}, \ldots, q_{ri}\}$ by decomposing and extracting based on $r$
4: **for** $r = 1$ to $R$ **do**
5:   **for** $q_r = 1$ to $Q_r$ **do**
6:     **if** RAG in $q_r$ **then**
7:       $K_{q_r} = \{k_{q_r1}, k_{q_r2}, \ldots, k_{q_ri}\} \leftarrow RAG(q_r, 10)$
8:       $K_{q_r} \leftarrow rerank(K_{q_r})$
9:       $K_{q_r} \leftarrow top\_k(K_{q_r}, 3)$
10:     **else**
11:       $K_{q_r} = \{k_{q_r1}, k_{q_r2}, \ldots, k_{q_ri}\} \leftarrow Tools(q_r, 3)$
12:     **end if**
13:   **end for**
14: **end for**
15: **for** $r = 1$ to $R$ **do**
16:   **Propose** $C = \{c_1, c_2, \ldots, c_i\}$ by reasoning confounding variables from $r$ and $K_r$
17: **end for**
18: $C' \leftarrow Ensemble(C)$
19: **Output:** $C'$
___

**Algorithm 2** Iterative Confidence Interval Estimation for Bias Reduction in Treatment Effect Inference
___
1: **Input:** $C'$, $S^{est}$, $S^{te}$
2: **Propose** $d \leftarrow ExpertDecision(C')$ by query expert whether variable C is clinically valid and relevant
3: **if** $d$ = reject **then**
4:   **return** $Agentic workflow$
5: **else**
6:   Apply restriction on $C'$ in $S$
7:   $B \leftarrow 64$
8:   **for** $b = 1$ to $B$ **do**
9:     $S_b^{est} \leftarrow Bagging(S^{est})$
10:     $T_b \leftarrow CausalTree(S_b^{est})$
11:   **end for**
12:   **for** $x_i = 1$ to $S^{te}$ **do**
13:     $\hat{\tau}_i \leftarrow \sum_{b=1}^{B} T_b(x_i)$
14:     $LowerCI_i = Q_{\alpha/2}(\hat{\tau}_{i1}, \ldots, \hat{\tau}_{iB})$
15:     $UpperCI_i = Q_{1-\alpha/2}(\hat{\tau}_{i1}, \ldots, \hat{\tau}_{iB})$
16:     $WidthCI_i = UpperCI_i - LowerCI_i$
17:   **end for**
18:   $threshold \leftarrow Mean(WidthCI_i)$
19:   **Filter** $S'$ by keep $x_i$ if $WidthCI_i > threshold$
20: **end if**
21: **Output:** $S'$, $WidthCI_i$
___

With $Q_r$ lead to reasoning, we need the knowledge to answer them. The agent first applies vector-based similarity ranking to retrieve a knowledge set $K_{q_r}$ from a domain-specific vector database. The retrieved $K_{q_r}$ are then reranked according to semantic proximity to the $q_r$, with the top three entries retained as the $K_{q_r}$. Suppose it is found that effective knowledge cannot be extracted from RAG. In that case, the agent transitions to tool-based retrieval. For example, it may access external databases like arXiv or PubMed to gather more knowledge to support the reasoning process.

Subsequently, the agent performs step-by-step causal reasoning, which generates a candidate set of confounders $C = \{c_1, c_2, \ldots, c_i\}$. Finally, ensemble learning techniques are applied to aggregate the $C$, resulting in a robust and stable set of confounding variables $C'$. The resulting $C'$ is then propagated to the next stage for expert review and integration into the unbiased estimation pipeline.

*C. unbiased estimation.*

To assess whether our estimated $\tau(X)$ is unbiased, we adopt confidence intervals $CI_i$ as the primary evaluation metric. As shown in Algorithm 2, the dataset $S$ is partitioned not only into training $S^{tr}$ and testing samples $S^{te}$, but also into a distinct estimation sample $S^{est}$ specifically reserved for validate the evaluation of the $\tau(X)$ between $\Pi$ and iterations. We generate multiple causal trees $\sum_{b=1}^{B} T_b$ to construct $CI_i$ for $\tau(X)$. Specifically, we train distinct $CausalTree(S_b^{est})$ on bootstrap subsamples $S_b^{est}$ of the $S^{est}$. By passing the $S^{te}$ through $\sum_{b=1}^{B} T_b$, we obtain the $CI_i$ of $\tau(X)$. The distribution of $CI_i$ is then used to compute the width of the confidence interval $WidthCI_i$ and the threshold $Mean(WidthCI_i)$. The $WidthCI_i$ reflects predictive stability: narrow intervals below the threshold indicate consistent and reliable results across bootstraps, whereas wide intervals above the threshold suggest potential prediction instability.

Next, we collect the samples that $WidthCI_i$ are above the threshold. These samples are identified as having unstable $\tau(X)$ and are believed to contain unaccounted confounding variables. By retaining these unstable samples $S'$, we enable their reallocation to more appropriate $\Pi$ in the next iteration of our analysis. Conversely, the samples whose $WidthCI_i$ fall below the threshold are considered stable because they have adequately controlled all confounding variables. These stable samples are filtered out since they no longer require further refinement.

Throughout the process, we iteratively identify confounding variables and collect $S'$ until the agent determines that no new confounding variables can be identified. At this stage, the iterative process terminates, resulting in a collection of causal trees, each tailored to a specific subgroup $\Pi$. Final subgroup $\ell$ assignments are obtained by tracing the decision path backward from the last iteration, ensuring that each sample is matched to the most appropriate $\ell$ based on its refinement history. For example, if no qualifying $\ell$ is found in the third iteration, we move to the second iteration, and so on. Since the number of confounding variables controlled increases with each iteration, a matching subgroup will always be found in the zeroth iteration. The $\tau(X)$ corresponding to this $\ell$ is taken as the final treatment effect estimate.

TABLE I.    CONFOUNDING VARIABLES DISCOVERED VIA LLM-BASED AGENTs

| Model | iteration1 | Iteration2 | Iteration3 |
|---|---|---|---|
| Med42-v2* | HTN, CHF, AF, CAD | DM | CVAD |
| palmyra-med* | HTN, CHF, AF, CAD | DM | CKD |
| Meditron* | DM, HTN, CHF, AF | CAD | — |
| OpenBioLLM* | HTN, CHF, AF, CAD | DM | CVAD |
| Med42-v2 | DM, HTN, CHF, CKD | COPDA | GOUT |
| Palmyra-Med | DM, HTN, COPDA, CKD | CHF | — |
| Meditron | DM, HTN, CHF, AF | CVAD | COPDA |
| OpenBioLLM | DM, HTN, CVAD, CKD | CAD | GOUT |

[a]Asterisk (*) indicates the use of BET as the treatment; models without * used ACE.

TABLE II.    PERFORMANCE COMPARISON ACROSS ALGORITHMS

| Algorithms | | BET | ACE |
|---|---|---|---|
| Causal Forest | | 0.211 | 0.250 |
| Generalized Random Forest | | 0.182 | 0.200 |
| **Our proposed with llama3-Med42** | 0 iteration | 0.260 | 0.246 |
| | 1st iteration | 0.148 | 0.167 |
| | 2st iteration | 0.139 | 0.148 |
| | 3st iteration | 0.133 | 0.141 |
| **Our proposed with palmyra-med** | 0 iteration | 0.260 | 0.246 |
| | 1st iteration | 0.147 | 0.137 |
| | 2st iteration | 0.138 | 0.132 |
| | 3st iteration | 0.133 | — |

[a]BET and ACE represent 95% CI widths; 'iter' denotes iteration.

## IV. EXPERIMENTS AND DISCUSSION

To demonstrate our proposed method, we employed data from Taiwan's National Health Insurance Research Database (NHIRD), a nationally representative longitudinal dataset comprising one million individuals from 2000 to 2008. Following the inclusion and exclusion criteria provided by Kaohsiung Medical University Chung-Ho Memorial Hospital (KMUH).

We derived a cohort focused on Acute Coronary Syndrome (ACS), including various categories of variables, such as demographic information, common comorbidities, and the selected medications for chronic diseases. We allocated 80% of the data to construct the training and estimation samples. ACS data aims to show whether the proposed approach can address unbiased estimation with the variability in medications, risk factors, and treatment strategies. In clinical causal inference, accurately identifying causal pathways from disease mechanisms to pharmacological interventions requires domain expertise. This task is tedious and time-consuming, involving the manual curation of hundreds of clinical rules and the consideration of dozens of interacting variables. To alleviate this burden and reduce reliance on manual expert input, we delegate the causal reasoning task to an LLM-based agent. Accordingly, we preload the vector database with authoritative ACS textbooks [20], [21], [22], [23] as knowledge sources and use the PubMed tool under our established Tool Use policy to facilitate the process of the agent on causal reasoning.

We adopt three open-source medical LLMs, including llama3-med42-70B [24], Palmyra-Med-70B-32k [25], Meditron-70B [26], and Llama3-OpenBioLLM-70B [27] for LLM-based agent experiments, primarily due to institutional data privacy requirements mandating execution on on-premise GPU infrastructure. Under this constraint, all experiments were conducted on a local computing server with four NVIDIA RTX 4500 Ada GPUs.

Table 1 compares confounding variable discovery across LLM models on the training sample. The termination criterion for the iterative process was defined as when the agent indicated that no additional confounding variables could be identified. This result aligns well with our goal of enabling the LLM-based agent to autonomously identify confounding variables, thus reducing the time required for clinical experts to evaluate each rule manually.

To assess the performance of our predictive approach, Table 2 presents the average widths of the 95\% confidence intervals on the testing set across three iterative stages for both models. This metric quantifies the precision of our predictions by reflecting the uncertainty in CATE estimates. We note the gradually narrowing CI widths of treatment effect estimation from iteration 0 to iteration 3. This iterative process reduces estimation uncertainty. Stable samples within the CI threshold in the training set are excluded. Unstable samples are re-evaluated by an LLM-based agent and a causal tree. Next, we compared our proposed approach with Causal Forest and Generalized Random Forest to the baseline performance. As shown in the upper portion of Table 2, traditional causal machine learning approaches produce considerably wider confidence intervals. This indicates higher levels of uncertainty in estimating treatment effects. Table 2 also suggested that our proposed approach enables the model to achieve narrower confidence intervals and produce interpretable subgroup rules while lowering experts' workload by replacing manual review with automated rule-based refinement.

Given the results in Tables 1 and Tables 2, stable samples within the CI threshold are excluded at each iteration, while unstable samples are recursively partitioned for re-evaluation. As shown in Figure 2, the number of unstable samples decreases over iterations, whereas the cumulative count of stable samples increases. Notably, 169 BET and 446 ACE samples exceeded the CI thresholds across all iterations. This persistent instability suggests the presence of unobserved confounders not captured by the current covariates. Such sample cases are typically difficult to detect, even for advanced ML, and often require expert screening. In our method, the iterative process identifies samples possibly confounded by unobserved confounders and reduces the human cost for domain experts.

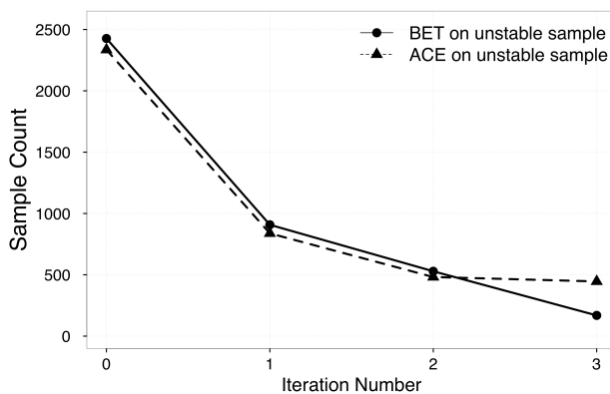

Fig. 2. Unstable sample data.

## V. CONCLUSION

This work proposes the use of an agent to improve treatment effect estimation through iterative identification of confounding variables. Our experimental results demonstrate that causal rule discovery via the agent achieves estimation performance comparable to standard machine learning models and identifies confounding variables that are otherwise difficult to detect. Our framework also provides rule-based subpopulation structures, which can be used in clinical decision-making and policy analysis. By offering an interpretable, rule-based approach, we present a compelling solution to the longstanding accuracy-interpretability trade-off, which follows that advancing the field of interpretable machine learning and causal inference.